# Døñ't Tòůčḫ Mý Dïäc̣r̥ītīc̄s

**Kyle Gorman**
CUNY Graduate Center
New York, NY
kgorman@gc.cuny.edu

**Yuval Pinter**
Ben-Gurion University of the Negev
Beer Sheva, Israel
uvp@cs.bgu.ac.il

## Abstract

The common practice of preprocessing text before feeding it into NLP models introduces many decision points which have unintended consequences on model performance. In this opinion piece, we focus on the handling of *diacritics* in texts originating in many languages and scripts. We demonstrate, through several case studies, the adverse effects of inconsistent encoding of diacritized characters and of removing diacritics altogether. We call on the community to adopt simple but necessary steps across all models and toolkits in order to improve handling of diacritized text and, by extension, increase equity in multilingual NLP.

## 1 Introduction

Virtually all natural language processing workflows begin with the ingestion of text data (with or without annotations). This data is usually not a random sample of written language, but rather has been sampled or filtered for language, script, quality, or relevance, and may also have been subject to various substitutions (e.g., case-folding, removing markup, or excising encoding errors). Yet preprocessing, as this process of text preparation is known, is generally regarded as more of a dark art than a topic for research, and as such has received minimal attention in the literature. While data preprocessing may not be the most important component of a speech or NLP system, any errors made at that stage are likely to propagate, and decisions made during preprocessing necessarily constrain what is possible downstream.

In this paper we focus our attention on the consequences of decisions made while preprocessing text with diacritics, a notion we define below. Using case studies, we show that failure to apply consistent Unicode normalization, which provides a canonical representation of text with diacritics, leads to degradation in downstream performance. We then show that stripping diacritics, common when pre-training large neural network language models (LLMs), also leads to degradation.

Our recommendations, then, are simple: text preprocessing regimens should apply a consistent Unicode normalization—any of the normalization forms will do—but in most cases, should not attempt to strip diacritics.

## 2 Defining diacritics

Written text consists of atomic units sometimes called *glyphs*. These glyphs act as the primary spatial units in text and their order mirrors the temporal ordering of the orthographically relevant linguistic units in the corresponding utterances (Sproat, 2000). Glyphs may also bear non-spacing marks appearing above, below, to the left or right, or even surrounding, the glyph; it is these marks we will call *diacritics*. It may be difficult to discern whether marks of these sorts are really "part" of a glyph or glyphs on their own, and judgments also seem to vary from language to language, reflecting *Sprachgefühl* or conventions learned in school.[1]

Reflecting this ambiguity, Unicode often provides multiple ways to encode diacritized glyphs. For example, an *e* with an acute accent can either be encoded either as a single character ⟨é⟩ (U+E9), or as an *e* (U+65) followed by a combining acute accent (U+301). Similarly, consider the Hindi word साड़ी 'sari'. In this word, one character is marked with a dot (a *nuqta*) underneath. With this dot the character is read as [ɽiː]; without it, it is read as [ɖiː]. In Unicode one can encode ⟨ड़⟩ either as a single precomposed character ड़ (U+95C) or as a se-quence of the undiacritized character (U+921) followed by a combining dot (U+93C).

---

[1] This uncertainty is not specific to diacritics. For instance, in Gajica, the Latin script used to write Serbo-Croatian, the digraphs ⟨dž⟩, ⟨lj⟩, and ⟨nj⟩ are conceptualized as single glyphs. Unicode, following earlier practices (in ISO 8859-2 and various vendor-specific encodings), provides unary codepoints for these glyphs (and their uppercase and titlecase variants), even though they are often rendered the same as two-character sequences and they decompose into character sequences in compatibility normalization forms.

|   |       |              |              |              |              |              |              |
|---|-------|--------------|--------------|--------------|--------------|--------------|--------------|
| a.| बाढ़:  | ब (U+92C)    | T (U+93E)   | ढ़ (U+95D)   |              |              |              |
|   | बाढ़:  | ब (U+92C)    | T (U+93E)   | ढ (U+922)    | ़ (U+93C)    |              |              |
|   |       |              |              |              |              |              |              |
| b.| عِدَّة: | ع (U+639)    | ِ (U+650)    | د (0x62F)    | َ (0x64E)    | ّ (U+651)    | ة (U+629)    |
|   | عِدَّة: | ع (U+639)    | ِ (U+650)    | د (0x62F)    | ّ (U+651)    | َ (0x64E)    | ة (U+629)    |

Table 1: Real-world Unicode canonicalization issues. (a): two different encodings of the Hindi word [bɑːɽʰ] 'flood', both found in the Hindi Dependency Treebank (Bhat et al., 2017; see §B). (b): two different encodings of the Arabic word [ʕiddah] 'number'; the former appears in the Prague Arabic Dependency Treebank (Smrz et al., 2008) in canonical order, and the latter occurs in the Arabic Broadcast News Transcripts (Maamouri et al., 2010) in a non-canonical order.

It is straightforward to apply Unicode normalization (see Appendix A for a brief tutorial) to convert between these two representations, but without normalization ⟨é⟩ and ⟨é⟩, and ⟨ड़⟩ and ⟨ड़⟩, are considered unequal by ordinary string comparison methods (e.g., the C standard library function strcmp, or the == operator in Python) despite the fact they are visually indistinguishable. At the same time, some characters that might naïvely appear to be diacritized forms of others are not regarded as such by Unicode. For example, the "belted L" ⟨ł⟩ used in Polish (among other languages) does not decompose into ⟨l⟩ and a diacritic as one might expect, nor does the "O with stroke" ⟨ø⟩ used in various languages of Scandinavia decompose into ⟨o⟩ and a diacritic. One interesting comparison is between the "square script" used to write Modern Hebrew and the (Perso-)Arabic script used for Modern Standard Arabic. In the former, consonant pointing (e.g., *dagesh lene* and the *sin*/*shin* dot) is considered optional and Unicode regards the diacritic as a separate character. In the latter, consonant pointing (e.g., the dots distinguishing *sīn* and *shīn*, and *ṣād* and *ḍād* respectively) is mandatory and the points are part of the glyph in all normalization forms.

Unicode also defines a canonical order for sequences of diacritics for those scripts in which a single glyph may bear multiple diacritics. For instance, in Arabic, in addition to the inherent consonant points, there are optional diacritics called *tashkīl*, including ones denoting quality of the following vowel (the *ḥarakāt*) and consonant gemination (*shaddah*). According to Unicode's canonical order, the *ḥarakāt* precede *shaddah*. Canonical order can be enforced by converting text to NFD or NFKD normalization forms.

## 3 The case for Unicode normalization

Applying Unicode normalization to text data enforces consistency with respect to two dimensions. First, it ensures consistency in whether diacritics are precomposed or decomposed. Secondly, in scripts where a glyph may bear multiple diacritics, it ensures that these diacritic sequences are in a consistent order. The normalization algorithms are deterministic, conceptually simple, computationally efficient, and available in the standard library of nearly all modern programming languages.[2] Yet even some professionally developed corpora lack consistent normalization; Table 1 provides two real-world illustrations.

It is not difficult to show that the failure to apply a consistent normalization would have negative consequences for NLP systems. For example, the example in panel (a) of Table 1, drawn from Hindi Dependency Treebank, was part of the CoNLL 2017 shared task on dependency parsing. Using the un-normalized Hindi data, we first replicate the system of Straka and Straková (2017), using their UDPipe 1.0 model and published hyperparameters: we obtain an labeled attachment score (LAS) of 87.09 on the test set. However, simply by applying Unicode normalization form NFKC—which composes the *nuqta* with its glyph—to the training and test data (and holding all other hyperparameters constant), we can improve the performance to a LAS of 87.38. While this improvement in LAS—0.29 absolute, 2.25 relative error reduction—may seem modest, it is available more or less for free: normalization brings a number of visually-identical distinct words into equivalence.[3] Similar issues would arise with the Arabic data in panel (b) in Table 1, were one to attempt to use data from the two data sources in a single system without first applying normalization.

---

[2]For C, C++, and Java, one is recommended to use the Unicode Consortium's open-source ICU library, available at https://icu.unicode.org/.
[3]Full details of this experiment are given in Appendix B.

| Language | Diacritized sentence | LLaMA | | mBERT | | XLM-RoBERTa | |
|---|---|---|---|---|---|---|---|
| | | raw | stripped | raw | stripped | raw | stripped |
| Spanish | Una olla de algo más vaca que carnero, salpicón las más noches, duelos y quebrantos los sábados, lantejas los viernes, algún palomino de añadidura los domingos, consumían las tres partes de su hacienda. | 63 | 62 | 61 | 60 | 57 | 56 |
| Arabic | صِرَٰطَ ٱلَّذِينَ أَنْعَمْتَ عَلَيْهِمْ غَيْرِ ٱلْمَغْضُوبِ عَلَيْهِمْ وَلَا ٱلضَّآلِّينَ | 62 | 28 | 47 | 12 | 77 | 26 |
| Hebrew | בְּבֹקֶר יוֹם רִאשׁוֹן הַזֶּבְרָה קָמָה, פָּשְׁטָה אֶת הַפִּיגָ׳מָה | 98 | 42 | 58 | 20 | 42 | 17 |

Table 2: LLM token counts of raw and diacritic-stripped text in Spanish, Arabic, and Hebrew; note that three words in stripped Arabic were mapped to [UNK] by mBERT.

## 4 The case for preserving diacritics

Many preprocessing regimens have no need to canonicalize diacritics for the simple reason that they remove them altogether. Diacritics recognized as such by Unicode can be stripped from text by converting to either of the decomposition normalization forms (NFD or NFKD), which causes the diacritics to be treated as separate characters, and then removing all characters in Unicode's "Mark, nonspacing" (Mn) category. This procedure is, for instance, used by the normalizers in Hugging Face's tokenizers library. Others may choose also to remove characters in the "Control" (C) or "Separator" (S) categories. An even more aggressive method for stripping pseudo-diacritics (like the Polish belted L) is proposed by Náplava et al. (2018). For instance, in their method the lower-case belted L ("Latin small letter L with stroke") is mapped onto lower-case L ("Latin small letter L") because the latter's full Unicode name is a proper prefix of the former's. The effect of stripping are evident in Table 2, where we present (non-cherry-picked) samples of tokenized text in Spanish, Arabic, and Hebrew. While the presence of Spanish diacritics minimally impacts the token count produced by the LLaMa (Touvron et al., 2023), multilingual BERT (Devlin et al., 2019), and XLM-RoBERTa (Liu et al., 2019) tokenizers, they struggle mightily with diacritized Arabic and Hebrew, requiring between two and four times as many tokens and often introducing token boundaries between glyph and the following diacritics.

**Not all languages are the same.** It is fairly obvious why one might choose to strip diacritics. First, if diacritics are inconsistently encoded, it might be better to simply remove them. However, as just discussed above, it is trivial to apply a consistent encoding to text. Secondly, there are many scripts which are only rarely written with diacritics. In Arabic and Hebrew, *tashkīl* and *niqqud* diacritics, respectively, are omitted except in certain pedagogical and religious materials. The same is true of the diaeresis used in Russian text to distinguish ⟨ë⟩ [jo] from ⟨e⟩ [je], or the acute accent used to indicate stress. In many other scripts, diacritics are ordinarily present but occasionally omitted due to haste or technical challenges. For example, the contrast between ⟨l, ł⟩ is of some importance in Polish, but it is easy to imagine a Polish speaker typing ⟨l⟩ in place of ⟨ł⟩ because of limitations in the available text entry system.[4]

In the presence of inconsistent diacritization, it might make sense to strip diacritics and thus use undiacritized word forms as equivalence classes for subsequent processing including tokenization. While there are likely some applications where this is a sensible decision, in many others it comes with measurable costs. Our colleagues in the social sciences (p.c.) report to us that off-the-shelf Hebrew speech recognition systems output undiacritized text. When they attempt to feed the resulting text into NLP analysis pipelines (e.g., POS taggers and dependency parsers), even state-of-the-art systems struggle with the ambiguity introduced by the absence of diacritics and often produce incorrect outputs that could be avoided if the recognizer produced diacritized text and the NLP pipelines accepted it. Modern Greek writing makes consistent use of acute accents to mark primary stress. Without these accents, ambiguities arise; e.g., νόμος 'law, ordinance' vs. νομός

---

[4]Older readers may have experienced similar issues composing a text in their native language on a "dumbphone", or writing emails on a desktop computer while abroad.

'county, district'. Yet GreekBERT (Koutsikakis et al., 2020), a pre-trained language model for Greek, is trained on text stripped of all diacritics.[5] Perhaps as a result, the model performs poorly when used for morphological analysis (Yakubov, 2024). Buhnila (2025) found that BioMistral (Labrak et al., 2024), when asked to define medical terms given in Romanian, not only performs better on undiacritized text (which is more common in everyday usage), but also tends to generate the diacritized form in a parenthetical remark preceding the definition. Kirov et al. (2024) study transliteration in twelve languages of South Asia. They experiment with a number of pre-trained language models, and find that one of them, mT5 (Xue et al., 2021), has poor vocabulary coverage in certain languages. This is because Malayalam and Telugu use the zero-width non-joiner character (U+200C) and Marathi and Sinhala use the zero-width joiner (U+200D), both to block inappropriate formation of conjunct (i.e., consonant cluster) characters, but these characters are absent from the mT5 vocabulary, presumably removed from the training data by an overzealous preprocessing routine.

**Inconsistent diacritics also have consequences.** Idiosyncratic appearance of diacritics in both training data and inference can also have unexpected effects. In a preliminary survey of production machine translation systems for Hebrew, we found that MarianNMT (Junczys-Dowmunt et al., 2018) performs inconsistently whether a word in a source sentence in Russian contains an accent mark or not. In some cases, such as пáдают 's/he falls', the presence of the accent mark in the input produces incorrect translation ('grow') but the translation is correct when unaccented. A similar verb, опáли 'they fell' produces the opposite effect: the incorrect translation occurs when the word is unaccented. Source sentences in Spanish introduced a different phenomenon: in a sentence with a feminine-marked subject, an unaccented (and incorrect) form of the past participle *burlándose* 'mocked' is translated as masculine in Hebrew. The accented form is translated correctly. Since MarianNMT's training sets retain Spanish diacritics, we hypothesize the unknown wordforms prevent the model from tracking grammatical agreement with nearby words. See Gonen et al. 2022 for similar observations.

---

[5]The BERT documentation reports that their original "uncased" checkpoints were also trained on stripped text.

## 5 Automatic diacritization

One way to mitigate the effects of stripped diacritics or inconsistent treatment of diacritics may come from *diacritization*, the task where undiacritized text is annotated with the correct marks. This is now a well-established task in NLP, particularly pertaining to consonantal scripts like those used for Arabic and Hebrew. However, this is something of an unsolved problem, as systems achieve just over 10% word error rate (WER) in Hebrew (Gershuni and Pinter, 2022). Similar WERs are reported for Arabic;[6] Náplava et al. (2018) report that WER exceeds 40% for Vietnamese. It may also seem that pre-trained neural language models would be quite effective of this task—in a zero-shot, few-shot, or fine-tuning scenario—even if they have been pre-trained on stripped text. While this certainly has been attempted, many of the state-of-the-art diacritization systems instead use randomly-initialized (rather than pre-trained) neural models. These models are robust, but are outmoded in most other NLP tasks, and one might be surprised to learn that a large amount of undiacritized text is less useful than relatively small amounts of in-domain diacritized data. We believe a major cause of LLMs' reduced ability to handle these tasks, demonstrated in Figure 1, is the fact that existing preprocessing ourtines prevent models from being exposed to diacritized training data.

## 6 Conclusion

In this opinion paper, we argue and provide evidence that decisions about how diacritics are treated during text preprocessing may have detrimental downstream effects on model performance. These effects can largely be mitigated by preserving diacritics and by using simple, deterministic methods for ensuring they are encoded consistently. We believe the examples we have reviewed only scratch the surface. Many other issues may reside unseen deep within large, black-box neural models where they are difficult to detect.

Mitigation of these effects, while not particularly hard to implement, does however require the attention and effort of many stakeholders. Let us consider the case of large pre-trained models, used either for feature extraction (i.e., as pre-trained encoders) or prompting. Given a pre-trained model, even one with detailed documentation (e.g., model cards) it is often difficult to determine what pre-processing steps were used to prepare data for the model. Furthermore, given such a model, it is not obvious how one might modify an existing checkpoint—or its tokenizer and vocabulary—of a pre-trained model to improve support for diacriticized text it has not yet been exposed to. Thus we are beholden to developers at institutions with sufficient resources to train such models and are in some sense stuck with the preprocessing decisions they have made.[7] It is for this reason that we appeal to the community at large, rather than quietly modeling what we have argued to be best practices. Developers of LLMs can increase their overall utility, particularly for languages other than English, simply by applying a consistent normalization and resisting the urge to strip diacritics.

---

[6]It is difficult to cite any one WER as state of the art since there are many different diacritized Arabic corpora—some proprietary—used for evaluating diacritization, and error rates vary widely. Methodical system comparison across a variety of publicly-available corpora is desperately needed.

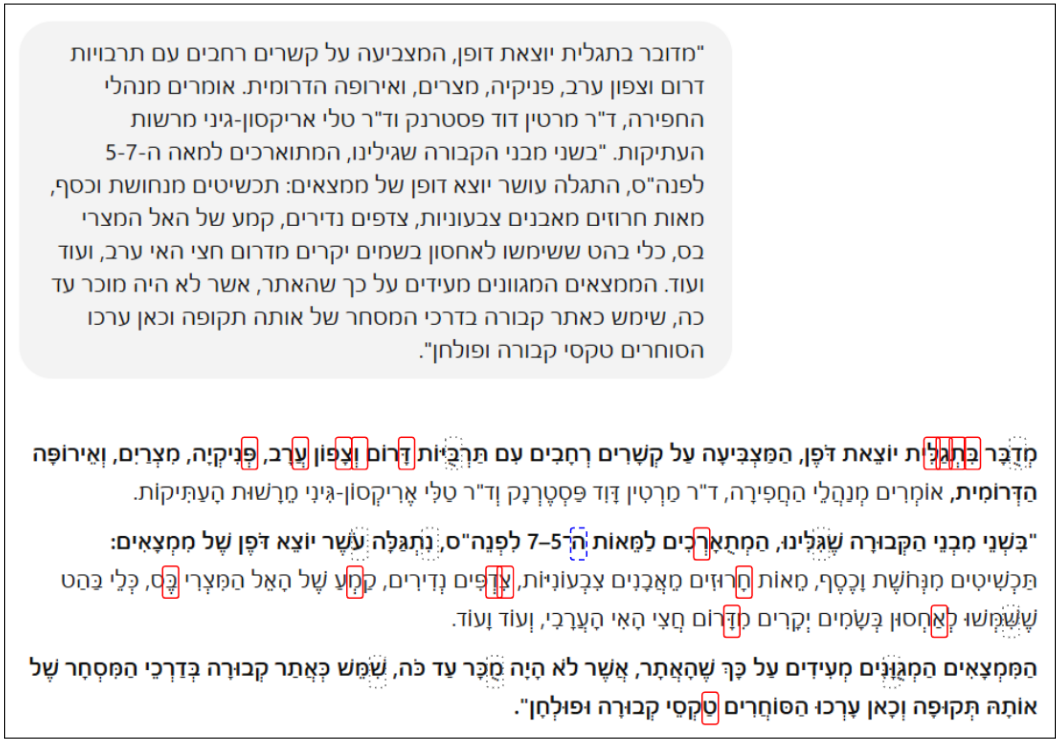

Figure 1: A sample of Hebrew text from a news website diacritized by a ChatGPT prompt ("turbo"), retrieved February 8, 2025. Errors marked in red; missing diacritics in dashed blue; and character edits in dotted gray.

We note that this preprocessing issue is orthogonal to the ongoing debate concerning the appropriate representation levels in language models. Byte-level and character-level models (e.g., Clark et al., 2022; Xue et al., 2022) are not immune to inconsistent encodings of diacritics, though it may be that they are better suited to represent inconsistent encoding than models with larger tokens. Vision-transformer processing of rendered text may help address invisible or near-invisible differences in rendered text (e.g., Lotz et al., 2023), but even here caution must be taken regarding any preprocessing performed before the rendering phase.

One might have expected that LLMs, given their impressive performance on a variety of tasks, would be robust to the use of diacritized inputs—or use in diacritization tasks—even when they themselves are not trained on diacritized text, but sadly this does not seem to be the case.

**Limitations**

The case studies above reflect the authors' experience and issues reported in the literature. We suspect that many issues similar to those we discuss have been encountered by others but have either gone unnoticed or unpublished. By bringing attention to these issues, we hope to encourage researchers to not only pay greater attention to diacritics and other text encoding issues, but also to encourage them to discuss these and other preprocessing decisions in the literature.

---

[7]For instance, Izsak et al. (2021), describe how to train a BERT-style model on an "academic budget", with a cluster of GPUs that, at time of writing, would cost roughly $4,000 US on the second-hand market. We suspect this budget is still well out of reach for many academic research groups.

## Acknowledgments

This research was supported by grant no. 2022215 from the United States–Israel Binational Science Foundation (BSF), Jerusalem, Israel. We thank Carinne Cherf for the translation examples, and the anonymous reviewers for an unusual amount of helpful feedback during review.

## A  Unicode normalization forms

Unicode provides four normalization forms, which are defined in terms of two types of equivalence:

- Codepoint sequences are said to be *canonically equivalent* if they have the same meaning and the same appearance when printed or displayed.

- Codepoint sequences are said to be *compatible* if they may have distinct appearances but the same meaning in certain contexts.

The normalization forms are computed as follows:

- NFC: decompose characters according to canonical equivalence, then recompose them according to canonical equivalence.

- NFD: decompose characters according to canonical equivalence, then order sequences of combining characters in canonical order.

- NFKC: decompose characters according to compatibility, then recompose them according to canonical equivalence.

- NFKD: decompose characters according to compatibility, then order sequences of combining characters in canonical order.

The Python standard library module `unicodedata` provides a function `normalize`, and this can be used to convert text strings between the four Unicode normalization forms. This function takes as its first argument the normalization form (e.g., `"NFD"`) and the string to be normalized as the second argument, returning the string in the desired normalization form.

## B  Hindi dependency parsing

Hindi experiments were were conducted using UDPipe v1.2.0 (https://github.com/ufal/udpipe, tag v1.2.0), word2vec (https://github.com/tmikolov/word2vec, commit 20c129a), and the Hindi Dependency Treebank (https://github.com/UniversalDependencies/UD_Hindi-HDTB/, commit 54c4c0f), targeting the "gold tokenization" subtask. Encoding inconsistencies with the Hindi treebank were reported by the authors to the maintainers, and this was marked fixed in commit da32dec (Dan Zeman, personal communication).